# Tracing the boundaries of materials in transparent vessels using computer vision

*Sagi Eppel*

Department of Materials Science and Engineering, Technion – Israel Institute of Technology, Haifa 32000, Israel.

E-mail: sagieppel@gmail.com

## Abstract

Visual recognition of material boundaries in transparent vessels is valuable for numerous applications. Such recognition is essential for estimation of fill-level, volume and phase-boundaries as well as for tracking of such chemical processes as precipitation, crystallization, condensation, evaporation and phase-separation. The problem of material boundary recognition in images is particularly complex for materials with non-flat surfaces, i.e., solids, powders and viscous fluids, in which the material interfaces have unpredictable shapes. This work demonstrates a general method for finding the boundaries of materials inside transparent containers in images. The method uses an image of the transparent vessel containing the material and the boundary of the vessel in this image. The recognition is based on the assumption that the material boundary appears in the image in the form of a curve (with various constraints) whose endpoints are both positioned on the vessel contour. The probability that a curve matches the material boundary in the image is evaluated using a cost function based on some image properties along this curve. Several image properties were examined as indicators for the material boundary. The optimal boundary curve was found using Dijkstra's algorithm. The method was successfully examined for recognition of various types of phase-boundaries, including liquid-air, solid-air and solid-liquid interfaces, as well as for various types of glassware containers from everyday life and the chemistry laboratory (i.e., bottles, beakers, flasks, jars, columns, vials and separation-funnels). In addition, the method can be easily extended to materials carried on top of carrier vessels (i.e., plates, spoons, spatulas).

# 1. Introduction

Visual recognition of material boundaries in transparent vessels is valuable in numerous fields and areas of applications. Such recognition is essential for analyzing properties such as liquid-level[1-31] and volume, as well as for identification of processes such as: precipitation, condensation, evaporation and phase-separation.[32-35,5] Recognition of the materials interface is particularly important in the chemistry laboratory[36] in which it is essential for various of analytical and synthetic methods such as liquid-liquid extraction,[15,12] column-chromatography, crystallization,[32] titration and distillation.[36,8] A computer-vision method that can perform this recognition could be used for analysis and automation of various such processes.[37,12,38-41]

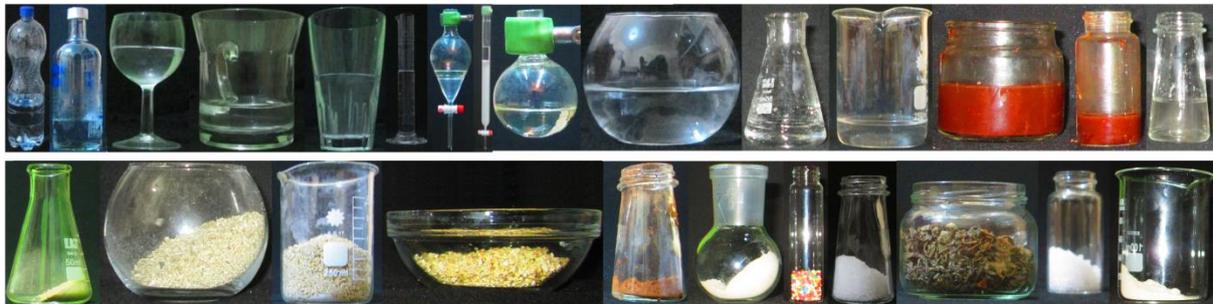

**Figure 1. Example of various fluids (Above) and solids (below) in transparent containers taken from everyday life and from the organic and analytical chemistry laboratories**

Computer-vision-based recognition of phase-boundaries for non-viscous fluids is usually achieved by assuming that the fluid has a flat surface (Figure 1 top).[21,42,30,5] In this case, the shape of the liquid surface in the image could be estimated as a straight line or an elliptic curve (Figure, 1 top).[5] The boundary of such surfaces could be found by scanning the image for all possible curves that correspond to the possible liquid surface shapes, as described in previous works.[5] However, materials that consist of viscous fluids, emulsions, foams, powders, granular particles or other types of solids will not necessarily display flat surfaces.[43-45] As a result, the boundaries of such materials in the image can take on a large number of possible shapes (Figure 1, bottom). In such cases, it is unrealistic to find the phase-boundary by simply scanning all possible surface shapes in the image. One approach for finding the boundaries of such materials is the use of standart image segmentation methods (i.e., graph cut, mean-shift) that divide the image into various regions according to certain image properties (i.e., color, intensity).[46-48] However, such segmentation approaches use only image analysis and completely ignore the physical constraints of the materials and vessels in the image.[49,50] One example of these physical constraints is that the material boundary must intersect the vessel boundary at both of its edges (Figure 2a). Another example of a physical restraint that occurs for all liquids, slurries, and granular/powder materials is that the slop of the material-interface is limited in steepness (Figures 1-2a). As a result of ignoring these restraints, standard image segmentation methods might falsely identify material boundaries with locations and shapes that are not physically possible for real materials (Figure 2b).[49,50] An image analysis method for finding the phase boundaries of materials in transparent containers must therefore use a combination of image properties and physical constraints. Such method will be examined in this work. The method is based on the assumption that the boundary of the material in the image must take the form of a curve that starts and ends at the transparent vessel contour and has limited slope steepness (Figure 2a). The general method is discussed in Section 2. Various filters and restraints that were applied to limit the recognition process to physically plausible shapes and positions are discussed in Sections 3-4. Several image properties that were examined as indicators for the material interface are discussed in Section 5. The results of the method for recognition of boundaries of various liquid and solid materials in various of transparent containers are discussed in Section 7.

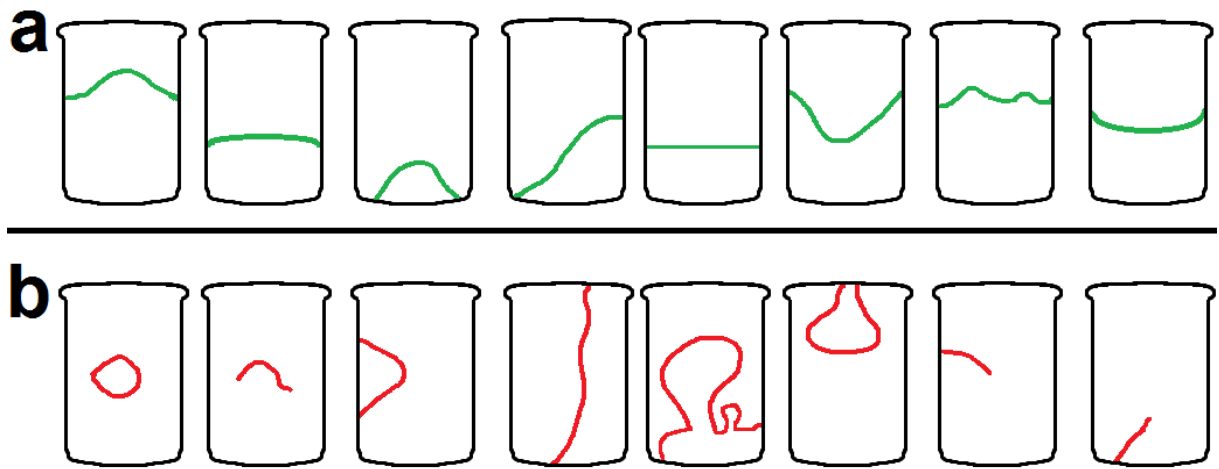

**Figure 2. Reasonable and unreasonable shapes for the boundaries of materials in transparent vessels (the vessel contour is marked in black): a) Examples of reasonable shapes of material boundaries in images, b) Examples of unreasonable shapes of material boundaries in images.**

## 2. General Method

Recognition of a material boundary in an image is carried out using an image of the vessel containing the material and the boundary of the vessel in this image (Figure 3a). The vessel boundary consists of all pixels in the image that are located on the vessel contour and could be represented as an array of coordinates or as a binary image (Figure 3a). Identification of the vessel boundary is performed beforehand using a template of the vessel shape or by segmenting the vessel area in the image from a uniform background.[51] The source code for finding the vessel boundary in the image using both methods is supplied as supporting materials. It is clear from Figures 1-2a that any curve that represents the material boundary in the image must start and end on pixels corresponding to the vessel contour (Figure 2). Based on this assumption, scanning of the image for the material boundary is achieved via the following five steps (Figure 3):

a) Scan every pair of pixels on the vessel boundary in the image and use them as endpoints of the boundary curve (Figure 3b).

b) For each pair of endpoints (a), use a physical constraint on the material interface to find the image region through which the path between this pair of endpoints can pass (green region, Figure 3c). Such physical constraints include a limited slope and steepness of the material surface and depend on the type of material and vessel examined (Sections 3-4).

c) Find a path (on the image) between the two endpoints used (a), that obey the physical constraints (b) and minimize a cost function that is based certain image properties along the path (Figure 3d). The image properties used for determining the path cost function should have an exceptionally high value at the image regions corresponding to the material boundary. Image properties that can act as indicators for material boundary include edges and sharp intensity or color changes normal to the path (Section 5). The values of these indicator properties along the path are used to calculate the path cost (Figure 3d), which represents the path curve's probability to overlap with the real material boundary in the image (Section 5). Finding the path with the lowest cost between the two endpoints is achieved using Dijkstra's algorithm.[52-54]

e) The path with the lowest cost out of all paths scanned is chosen to represent the material boundary in the image (Figure 3e). Similarly, for cases in which the image contains more than one material phase (i.e., a solid immersed in liquid or phase-separating liquids), it is possible to choose the two best curves as the phase-boundaries. If the number of the material phases in the vessel is unknown, it is possible to choose a threshold cost and accept any curve with a cost lower than this threshold as a phase boundary. One method of picking this threshold is the cost of the best path found in the scan multiplied by a given constant (as shown in previous works[5]).

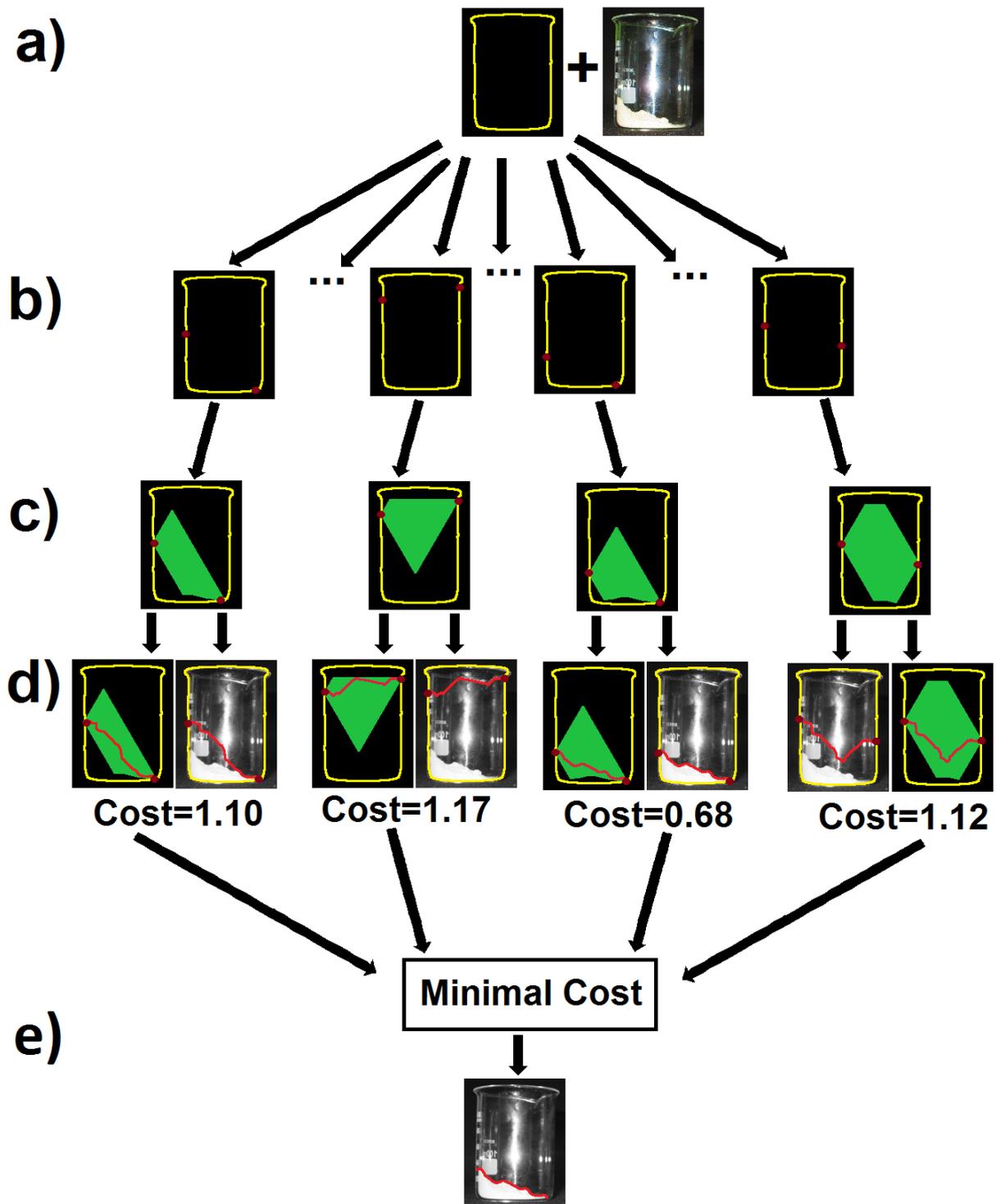

**Figure 3.** General method for finding the material boundary in the image: a) Receive the image of the material in a transparent vessel and the boundary of the vessel in this image; b) Scan every pair of points on the vessel boundary in the image; c) Use a physical constraint to limit the image regions through which the phase-boundary curve between this pair of points can pass; d) Find the optimal path (on the image) between this pair of points according to a selected cost function based on an image property that act as an indicator for the material boundary; e) The path with the lowest cost in the entire image is chosen as the material boundary.

The simplified Matlab code for the above method is given in the appendix. The steps described above represent a simplified description of the algorithm, which runs relatively slowly due the need to scan every pair of points on the vessel contour (Figure 3a). Practical implementation of this method does not demand calculation of the best path/curve between every pair of vessel contour points separately because Dijkstra's algorithm (used for finding the best path between two points) finds all optimal paths/curves leading from a given point simultaneously in one iteration (see appendix for more details).

## 3. Constraints of endpoints of the phase boundary curve

### 3.1. Path steepness and angle constraint

In an image, the boundary of a material in a glass vessel must take the form of a curve that starts and ends on the vessel outline (Figure 2a). However, not every pair of vessel contour points represents reasonable endpoints of the boundary curve (Figure 2b). A major limitation is the steepness of the slope of the line between the path's starting and ending points. For example, in liquids, the endpoints of the material boundary in the image are located mostly at the same height (Figure 1, top). For powders, slurries, foams, and granular solids, the slop of the material surface is also limited by the physical properties of the material (Figure 1, bottom). Limitation of the general slope of the boundary was achieved by ignoring paths for which the slope of the line between the path's starting point and ending points exceeds a selected threshold ($\theta$, Figure 4). In other words, all pairs of points on the vessel contour that form a line with a slope that exceeds a minimal steepness were ignored during the scan (Figure 4). In this work, the angle of the line between the path's starting and ending points ($\theta$, Figure 4) was limited to 55° degrees for all cases, although a much smaller angle could be used for liquids (0-5°).

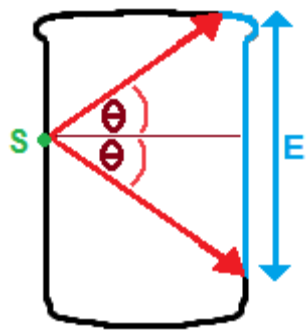

**Figure 4: To limit slope of the material boundary, the angle of the vector between the boundary curve's starting point (S) and end point (E) must not exceed a given threshold angle θ. For a given starting point (S), only ending at points in the range E will be scanned.**

## 3.2. Minimal length constraint

Ignoring paths that are too short is also necessary to assure reliable results. Short paths in which the horizontal distance between the path's start and end points is small could contain notably few pixels and focus on a limited region of the image. As such, these paths produce an unreliable path cost that can lead to false recognition. In addition, narrow vessel regions correspond to corks and funnels which can disturb recognition. To avoid this problem, paths in which the horizontal distance between the endpoints in the image is smaller than a selected fraction of the vessel's average width where ignored.

## 4. Physical and constraint on boundary curve shape and location

The boundary curve of the material in the vessel is limited not only by its starting and ending points (Section 3) but also by the local slope and the location of every point on this curve. The most obvious constraint is that the material boundary path most be contained entirely inside the vessel region in the image (Figures 1 and 2). Other examples of constraints include limits on the material-boundary slope steepness for liquid and powder materials. The slope of the interface for liquids as well as powders and granular materials is usually limited by physical constraints and is unlikely to reach high steepness values both locally and globally (Figure 1). Additional restraints on the path result from the type of transparent vessel used and are applied to discourage or prevent the propagation of paths in image areas that are likely to cause false recognition. Examples of such image regions are pixels in close proximity to the vessel boundary as well as areas in the image corresponding to vessel corks and funnels. Local constraints were applied to the phase-boundary curve using three different methods that are described in Sections 4.1-4.3.

### 4.1. Constraining the local propagation direction and slope of the path

Identification of the best boundary curve between a given pair of points in the image was carried out using the Dijkstra's algorithms to explore the best path between two points on a grid.[53,54,52] This algorithm is applied by propagating the path in a pixel-by-pixel manner from a given starting position (**S**) to the path's ending position (**E**, Figure 5c-d). Limitation of the path's local slope was achieved by constraining the direction of the path propagation in each step of the path exploration (Figure 5a). The path propagation was limited to movement from left to right and vertical movement, as shown in Figure 5a. Propagation from right to left (Figure 5b) was forbidden, which limits the path to a maximal slope angle of 90° (Figure 5c) and prevents back propagation and unreasonable paths (Figure 5d-2b).

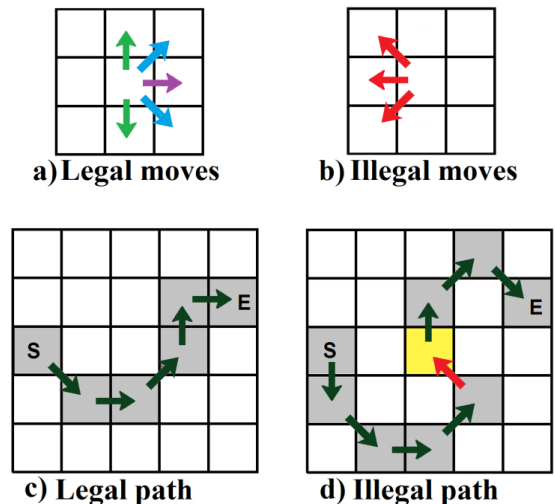

Figure 5. a) Legal directions for movement in the boundary path exploration include all moves to neighboring pixels from left to right and vertical moves; b) Illegal moves in path propagation include all moves from right to left; c) Example of a legal path from S to E (only left to right and vertical moves); d) An illegal path from pixel S to E, due to illegal backward move.

### 4.1.1. Limiting the path slope by limiting the number of sequential vertical moves

The path's local slope could be further restrained by limiting the number of sequential vertical propagation steps that could be performed in a row (Figure 6) or, in other words, limiting the length of vertical sections within the path curve. Limiting the length of the vertical sections in the path to one or two pixels limits the path slope angle to 63° and 72°, respectively (Figure 6b, c). Banning vertical path propagation altogether limits the local path angle to 45° (Figure 6a). In this work, the number of vertical steps that can be performed in a row was limited to three, which limits the angle of the path local slope to 76° (Figure 6d). An additional method that was used to discourage paths with high steepness is the addition of a penalty cost to each vertical move in the path (increasing the cost of vertical moves by 20%).

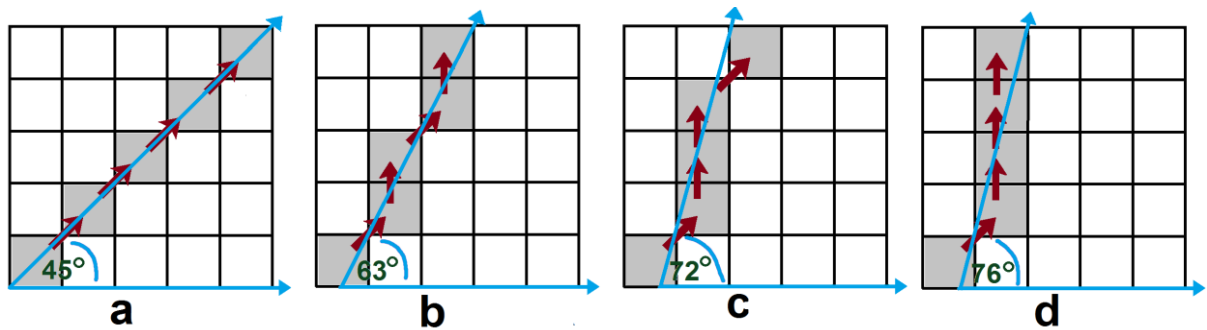

Figure 6. Limiting the path local slope by limiting the sequential number of vertical propagation steps: a) Forbidding vertical moves limits the angle to 45°; b) Limiting the number of sequential vertical moves to one limits the slope to 63°; c) A limit of two sequential vertical moves limits the path angle to 72°; d) A limit of three sequential vertical moves in an a row limits the path slope angle to 76°.

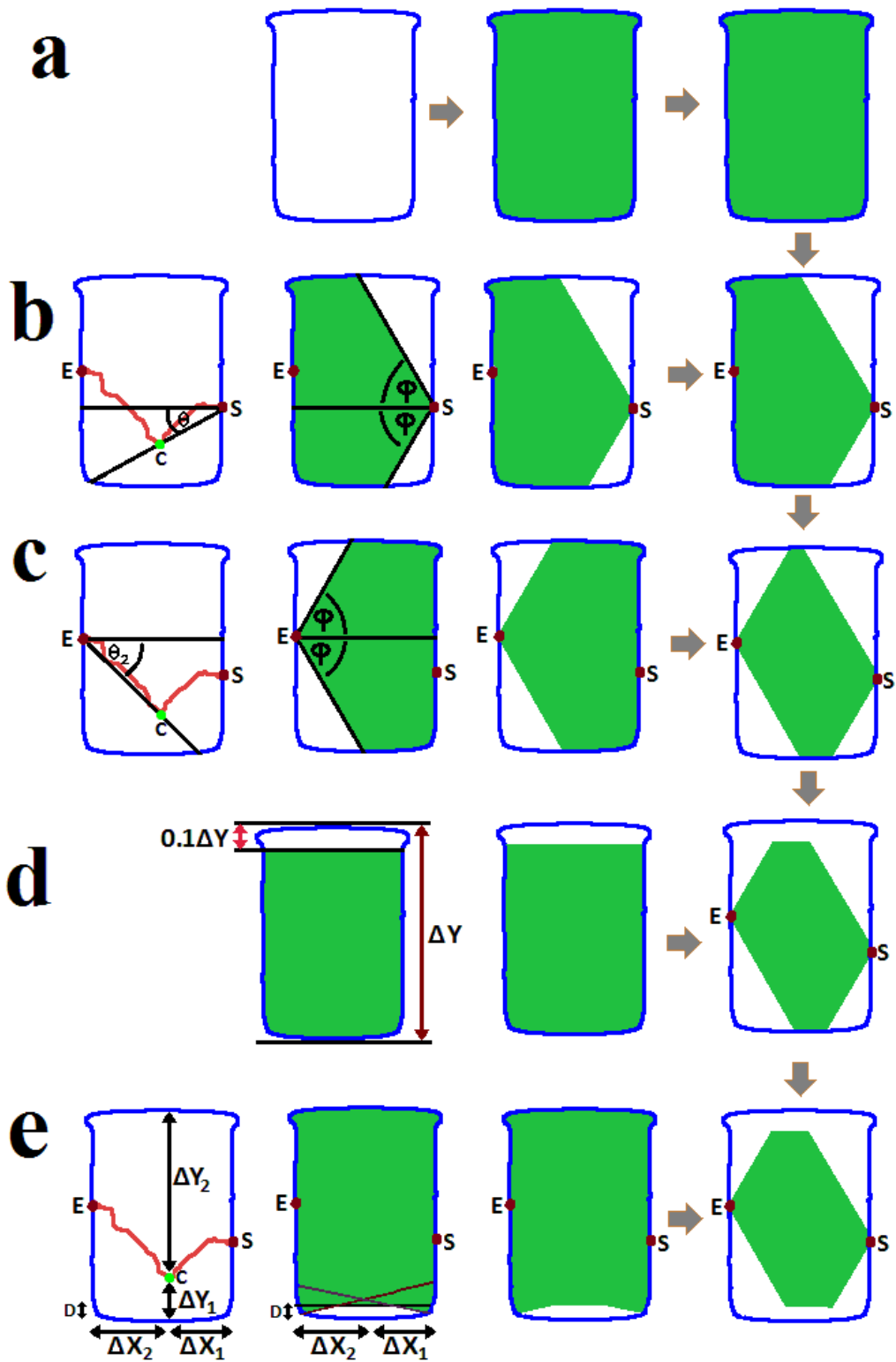

Figure 7. Legal image regions for a path between point S and point E (the legal region is marked in green, and the vessel boundary is marked in blue): a) The material boundary must be located within the vessel region of the image; b, c) Slope filter: The angle ($\theta$) of the vector between any point on the path (C) and the path's endpoints (E, S) must not exceed a given threshold ($\varphi$); d) Top/bottom filter: The path is not allowed to propagate in the top 10% of the vessel; e) Flat-path filter: The vertical distances ($\Delta Y_1$, $\Delta Y_2$) between every point on the path (C) and the vessel boundary must exceed a certain fraction of the minimal horizontal distances ($\Delta X_1$, $\Delta X_2$) between the point (C) and the path edge points (E, S) or a given minimal constant distance D; Hence, $\Delta Y_{1,2} >$ min($\Delta X_1/4$, $\Delta X_2/4$, D).

## 4.2. Constraining the image regions in which the path can propagate

Limiting the image region through which a path between a given pair of pixels is allowed to propagate (Figure 3c-d) is yet another method used to apply physical constraints on the material boundary. This process was carried out by predefining pixels in the image through which the path between two points could pass (green, Figure 3c-d). Initially the group of legal pixels contains all pixels within the vessel region of the image (Figure 7a). The addition of any further restraints removes more pixels from the list of legal pixels through which the path can pass (Figure 7b-e). The constraints that were applied using this method are discussed in Sections 4.2.1-4.2.3 and shown in Figure 7.

### 4.2.1. Path slope and angle limitation

As mentioned previously, the slope of the interface for liquids as well as powder and granular solids is usually limited by physical constraints and is unlikely to reach high steepness values (Figures 1 and 2). This constraint was previously applied at two levels (Sections 3.1 and 4.1). Another mode in which the path slope is restrained is by limiting the angle of the line between the path's start point (**S**) and any point (**C**) on the path (**SC**, Figure 7b). Hence, any pixel (**C**) for which the vector to the path's start point (**SC**, Figure 7b) exceeds a specific angle ($\varphi$) is removed from the group of legal pixels through which the path can pass (Figure 7b). The path contains two endpoints, and as a result, the constraint was applied twice, i.e., once for each endpoint (**S**, **E**, Figure 7b-c).

### 4.2.2. Bottom floor filter for irrelevant region of the vessel

Many glass vessels (i.e., bottles, jars and separatory funnels) contain corks, funnels, valves or other functional parts. These parts usually have strong edges that can be mistakenly identified as material boundaries in the image (Figure 8). Ignoring such features in the scan is therefore necessary to avoid false recognition. This process is carried out by simply removing all of the pixels in regions of the image corresponding to these parts (corks, funnels and valves) from the list of positions through which the boundary curve can pass (Figure 7d). In most cases, these parts are located in either the bottom or top areas of the vessel (Figure 8). Ignoring image regions in the vessel top and bottom is a general method for addressing these cases (Figure 8). To apply this process, every pixel in the image for which the vertical distance from the vessel top or bottom is smaller than a given threshold distance was removed from the list of legal pixels (Figure 7d).

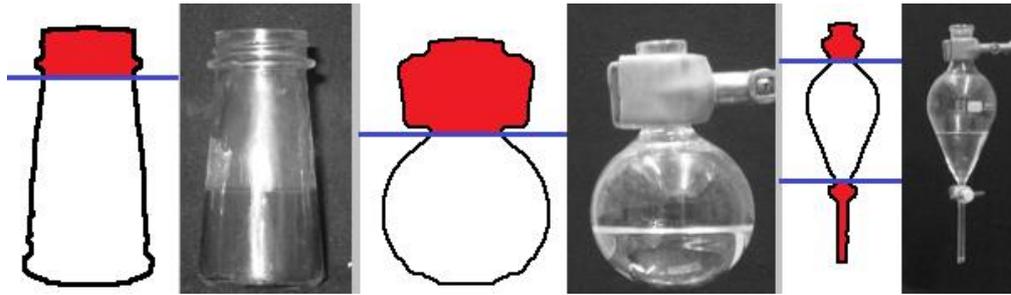

**Figure 8. Areas at the top and bottom of the vessel region of the image (red) often correspond to corks, funnels and valves and are therefore ignored.**

### 4.3. Preventing false recognition for paths along the vessel boundary

Propagation of the path along the boundaries of the vessel in the image is another major problem that could cause false recognition. The area in the image close to the boundary of the material container vessel is often characterized by strong edges (Figure 9). These properties, are also used as indicators for the material interface and thus can cause false recognition for paths along the vessel boundaries. Two methods to address this problem are described in Sections 4.3.1 and 4.3.2.

### 4.3.1. Flat path restraints

The flat path restraint was added to prevent flat paths along the vessel top and bottom boundaries. The vertical distance ($\Delta Y_{1/2}$, Figure 7e) between any point (**C**) on the path and the vessel boundary in the image must exceed 25% of this point's (**C**) horizontal distance ($\Delta X_{1/2}$, Figure 7e) to the path's closest edge points (**S**, **E**, Figure 7e) or a selected minimal distance $D$. Hence $\Delta Y_{1,2} > \min(\Delta X_1/4, \Delta X_2/4, D)$.

### 4.3.2. Penalty zone surrounding the vessel boundary

A second method used to discourage false recognition of paths along vessel boundaries is the creation a penalty zone in image regions near the vessel boundary (Figure 9). The cost of path propagation in this region of the image will be tripled. As a result, the paths along the vessel boundary will have higher costs and will be discouraged (Figure 9).

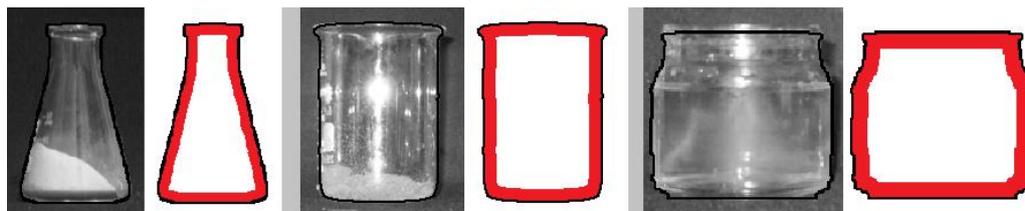

**Figure 9. Penalty zone: Image areas near the vessel boundary (marked black) are more likely to cause false recognition, and therefore, all pixels with a certain distance of the vessel boundary (red) are considered located in a penalty zone where the cost of the path is tripled.**

# 5. Using the cost function to evaluate the match between the curve and the material boundary in the image

Evaluation of the overlap between a given curve and the boundary of the material in the image is achieved by evaluating a specific property of the image surrounding each pixel in the path. Ideally, the image property chosen as an indicator for the material boundary should have high value in image regions that correspond to the material boundary and low value in other regions. The boundaries of material in an image are usually characterized by strong edges and intensity (or color) change perpendicular to the boundary curve. Therefore, these properties were examined as indicators for the phase-boundary.

## 5.1. Cost function and correlation between the pixel and the material boundary

Numerical evaluation of the overlap of a given pixel in the path with the material boundary in the image was achieved using a cost function.[52,54] The cost function takes the value of an image property in a pixel ($P$) and uses it to calculate the cost of the path at this pixel. The lower the cost, the higher the probability will be that the point overlaps with the boundary of the material in the image. The value of the image property used as an indicator must have a positive correlation with the material boundary. Therefore, the simplest form of a cost function is: $Cost(P) = C \cdot Prop(P)$, where $Cost(P)$ is the value of the cost function at point $P$, $Prop(P)$ is the value of the image property used as indicator in point $P$, and $C$ is a constant. As discussed in Section 4.3.2, to avoid paths that follow the vessel boundary, if a pixel is located at a given minimal distance from the vessel boundary in the image (Figure 9), the cost of the path in this pixel is tripled.

## 5.2. Evaluating the path cost

Given the cost function for each pixel in a path, it is possible to calculate the complete path cost as the sum of the costs of all points in this path. The lower the path cost, the higher the probability will be that the path corresponds to the material boundary in the image. The problem with this approach is that long paths with many points will automatically accrue higher costs compared with short paths with few points (Figure 10). As a result, the material boundaries in wider regions of the vessel are more likely to be missed, whereas paths in narrow regions of the vessel are more likely to be falsely identified as boundaries (Figure 10). To solve this problem, the path cost was normalized using the path horizontal distance ($ΔX$,

Figure 10). Hence, the path cost was calculated as the sum of the costs of all pixels in the path divided by the horizontal distance (*ΔX*) between the path edge points (Figure 10).

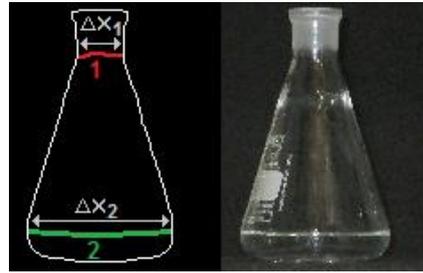

**Figure 10. To avoid bias toward short paths, the cost of each path (1, 2) is normalized by dividing by the horizontal distance (ΔX$_1$, ΔX$_2$) between the path edges.**

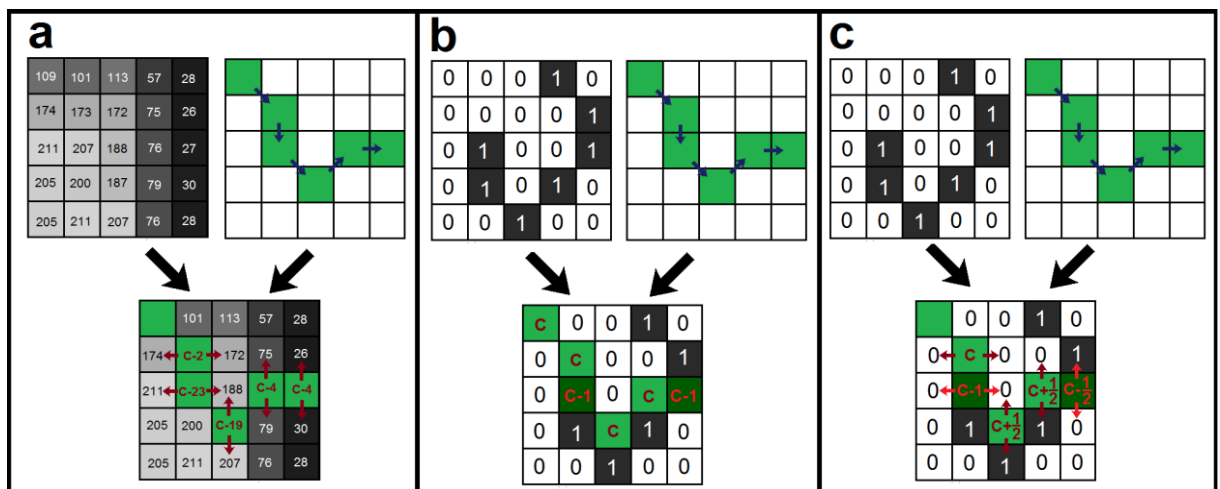

**Figure 11. Three cost functions based on grayscale images (a) and edge images (b, c). For all panels, the upper left matrix represents an image fragment and the upper right matrix represents the path on this image fragment. The lower central matrix shows the path overlay on the image with the cost of each point marked in red: a) The cost of pixels on the path is a constant (C) minus the difference between the intensities of pixels on both sides of the path; b) The cost is C (constant) for pixels not located on the edges and C-1 for pixels located on the edges; c) The cost is a constant C minus the difference between the edge intensity of the pixel (0/1) and the average edge density of its surrounding.**

### 5.3. Intensity change as an indicator for the material boundary in an image

Boundaries of features in images are usually characterized by sharp intensity or color changes across the feature boundary curve (Figure 1). Use of the change in intensity or color normal to the path as an indicator for the path correspondence with the material boundary in the image could be applied easily by taking the cost of a point **P** in the path as $\mathrm{Cost}(\mathbf{P}) = C - \Delta I(\mathbf{P})$, where $C$ is a constant, and $\Delta I(\mathbf{P})$ is the size of the intensity change perpendicular to the path in point **P** (Figure 11a). The problem with using the intensity change as a boundary indicator is that areas with strong illumination and high brightness/intensity can display stronger intensity changes regardless of whether they contain real boundaries. To solve this problem, it is possible to use the relative intensity change normal to the curve as an indicator for the phase boundary in the image.[5] This process is implemented by taking the cost of point **P** in the path as: $Cost(\mathbf{P})=C\text{-}\Delta I(\mathbf{P})/I(\mathbf{P})$ where **C** is a constant, I(**P**) is the intensity at point **P**, and ΔI(**P**) is the intensity change normal to the path at point **P**. The relative intensity change is a better indicator for liquid boundaries, whereas intensity change is a better indicator for boundaries of solid materials.

## 5.4. Edge-based indicators for the material boundary

Edges are the best indicators for the boundary of features in images. Identification of edges in images has been explored extensively, and every image-processing package contains a built-in implementation of edge detector.[55-57] Edge detectors are functions that receive an image and return an edge image in the size of the original image (Figure 12). Edge images are usually binary images in which every pixel in a location that corresponds to an edge in the original image is marked 1 (white, Figure 12), and all other pixels have values of zero (black, Figure 12).

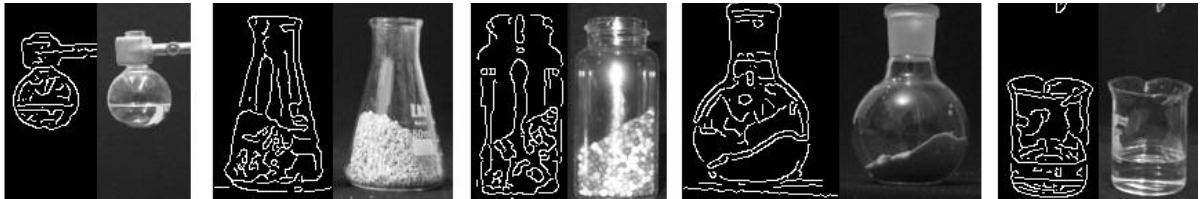

**Figure 12. Examples of images and their corresponding edge images (canny edge detector).**

### 5.4.1. Cost function for path correlation with edges

A simple method for using edges to trace material boundaries is to create a cost function that assigns low costs for points on the path that overlap with edges (Figure 11b). This process can be easily implemented by taking the cost of a point **P** in the path as $Cost(\mathbf{P})=C\text{-}Edge(\mathbf{P})$, where $C$ is constant, and Edge(**P**) is the value of pixel **P** in the edge image (1 if pixel **P** corresponds to edge and 0 otherwise, Figure 11b). This method gives good results; however, the drawback is that areas in the image with dense features also have a high density of edges that can cause false recognitions (Figure 12). One method for avoiding false recognition in image areas with dense edges is to normalize the cost of the path by the edge density of its close surroundings. Paths surrounded by areas with high edge density will receive lower costs and will be less likely to be accepted as the phase-boundary. This process was implemented using the difference between the edge density on the pixel (**P**) of the path and its surrounding pixels on both sides of the path (Figure 11c). Hence, the cost of a point **P** on the path was calculated as Cost(**P**)=$C$-[Edge(**P**)-Edge_Density(**P**)], where $C$ is constant, and Edge(**P**) is the value of pixel **P** in the edge image (1 if this pixel corresponds to edge and 0 otherwise). The Edge_density(**P**) is the average edge density of the surrounding of pixel **P** (the average edge density on both sides of the curve next to point **P**, Figure 11c). This cost function reduces false recognition in areas with strong texture and dense edges and yields better recognition accuracy (Tables 1-2).

## 5.5. Finding the optimal path between two points using the Dijkstra's algorithm

Given a starting point and an ending point for a path (Figure 3b) and a cost function (Section 5), the only challenge that remains is to find the path between the starting and ending points with the minimal cost. This is a simple optimization problem and can easily be solved using the well-explored Dijkstra's algorithm.[53,54,52] This algorithm gives the optimal (global) best path in polynomial time. The source code and documentation with implementation of the algorithm for this problem are given in the supporting material.

## 6. Experimental

The recognition method was tested using a set of 251 images containing transparent vessels with various materials. This set contained 151 images of vessels containing solid materials and 101 images of vessels containing fluids. The glass containers used in the image included ordinary glass vessels (i.e., jars, bottles and cups) as well as glassware used for analytical chemistry and organic synthesis (i.e., beakers, chromatographic columns, separatory funnels, Erlenmeyer flasks, round-bottom flasks, and vials).[36] The solids in the vessels included various powders, grained materials, and dry leaves with various particle sizes and morphologies; certain solids were immersed in liquids to examine the recognition of liquid-solid interfaces. The liquids used in the images included water, oil, silica slurries, and various organic solvents (DMF, hexane, etc.). All pictures were taken using a uniform, black, and smooth curtain fabric with no folds as a tablecloth and background. The areas belonging to the vessel in the image were recognized using template matching or by extracting the vessel region from the uniform background based on its symmetry.[58] The Matlab source codes for all methods and documantation are supplied in the supporting material.

## 6.1. Evaluating the test results

The search for the material boundary in the image was carried out based on the methods described in the previous sections and the conditions in Section 6. Only the path with the lowest cost was chosen for each image. The best path found for each image was manually compared with the boundary of the material in this image and assigned one of four levels of matching: 1) Full match: The path found and the material boundary in the image completely overlap; 2) Good match: The path found and the phase boundary in the image overlap nearly completely with minor deviations (approximately 90% or more); 3) Partial match: The path found overlaps with the material boundary in most areas but deviates from it in major regions; 4) Low match: The path found mostly misses the material boundary in the image but overlaps with it in a few regions; 5) Complete miss: The path does not overlap with the material interface in any region. For true/false evaluation of the results, only the perfect and good matches (Cases 1 and 2) were considered as true recognitions (Figure 13). Partial and low matches (Cases 3-5) were considered as false recognitions (Figure 14). The results are given in Tables 1-2.

**Table 1: Result of material boundary recognition for images of solid/powder materials.**

| Case[a] | True match[b] | False match[b] | Perfect match[2] | Small deviation[b] | Minority deviation[b] | Majority missed[b] | Complete missed[b] | Indicator description[a] |
|---|---|---|---|---|---|---|---|---|
| 1 | 80% | 20% | 68% | 12% | 5% | 6% | 8% | Intensity change normal to curve[c] |
| 2 | 74% | 26% | 66% | 8% | 8% | 4% | 14% | Relative intensity change normal to curve[c] |
| 3 | 88% | 12% | 81% | 7% | 3% | 2% | 8% | Difference between edge density on and around the curve[d] |
| 4 | 83% | 17% | 77% | 6% | 4% | 2% | 11% | Edge density on curve[d] |

[a] Each row gives the result of a different cost function/indicator. The image property used by the indicator is given in the indicator description column and is explained in Section 5.
[b] See Section 6.1.  [c] See Section 5.3.  [d] See Section 5.4.

**Table 2: Result of material boundary recognition for images of liquid materials.**

| Case[a] | True match[b] | False match[b] | Perfect match[2] | Small deviation[b] | Minority deviation[b] | Majority missed[b] | Complete missed[b] | Indicator description[a] |
|---|---|---|---|---|---|---|---|---|
| 1 | 58% | 42% | 47% | 12% | 10% | 5% | 27% | Intensity change normal to curve[c] |
| 2 | 67% | 33% | 55% | 12% | 7% | 5% | 21% | Relative intensity change normal to curve[c] |
| 3 | 82% | 18% | 76% | 6% | 2% | 1% | 15% | Difference between edge density on and around the curve[d] |
| 4 | 75% | 25% | 67% | 8% | 2% | 3% | 20% | Edge density on curve[d] |

[1-4] All notes are identical to the notes in Table 1.

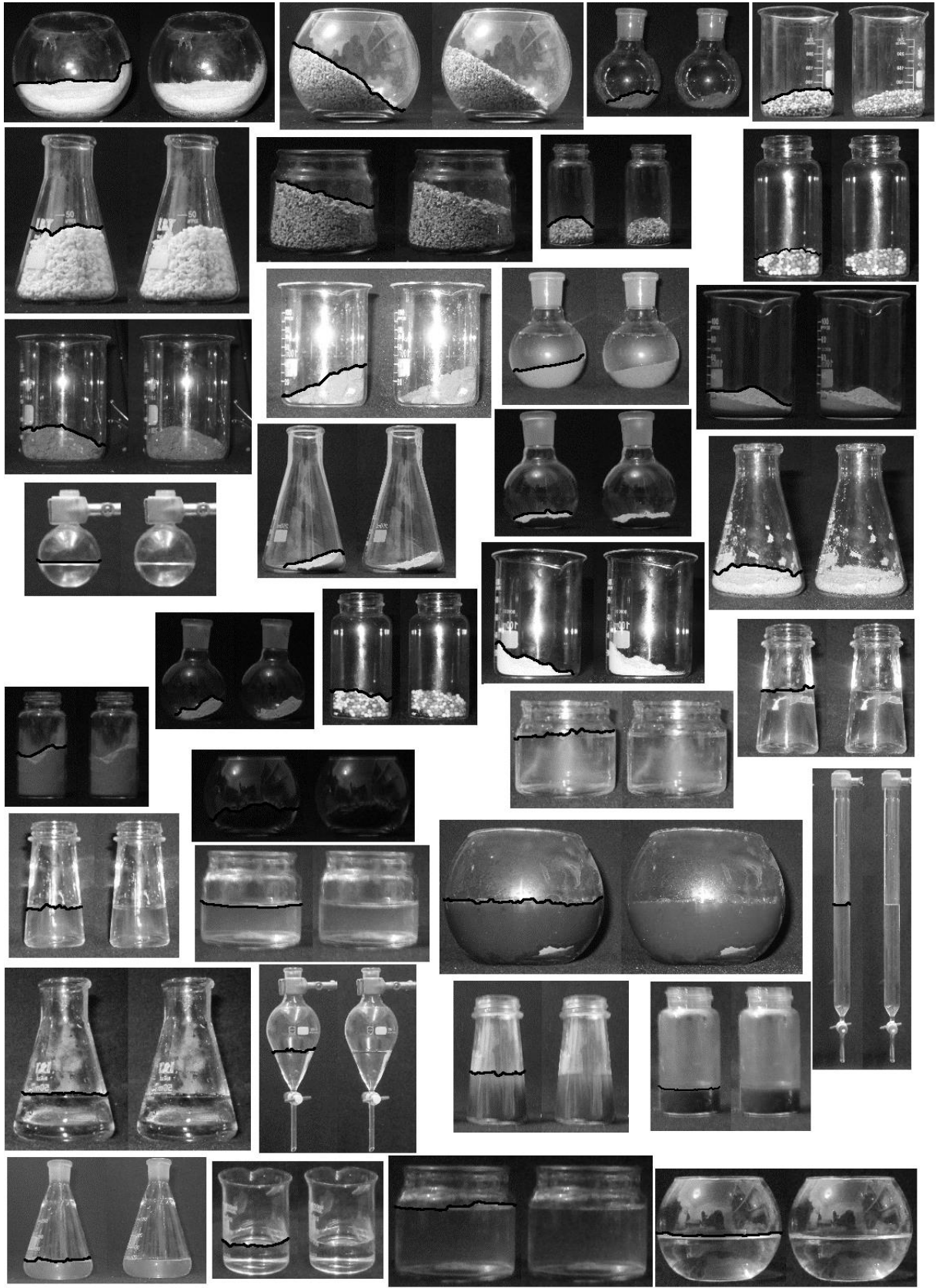

**Figure 13.** Examples of true matches between the path found and the material boundary in the image. Based on method 3, Tables 1-2

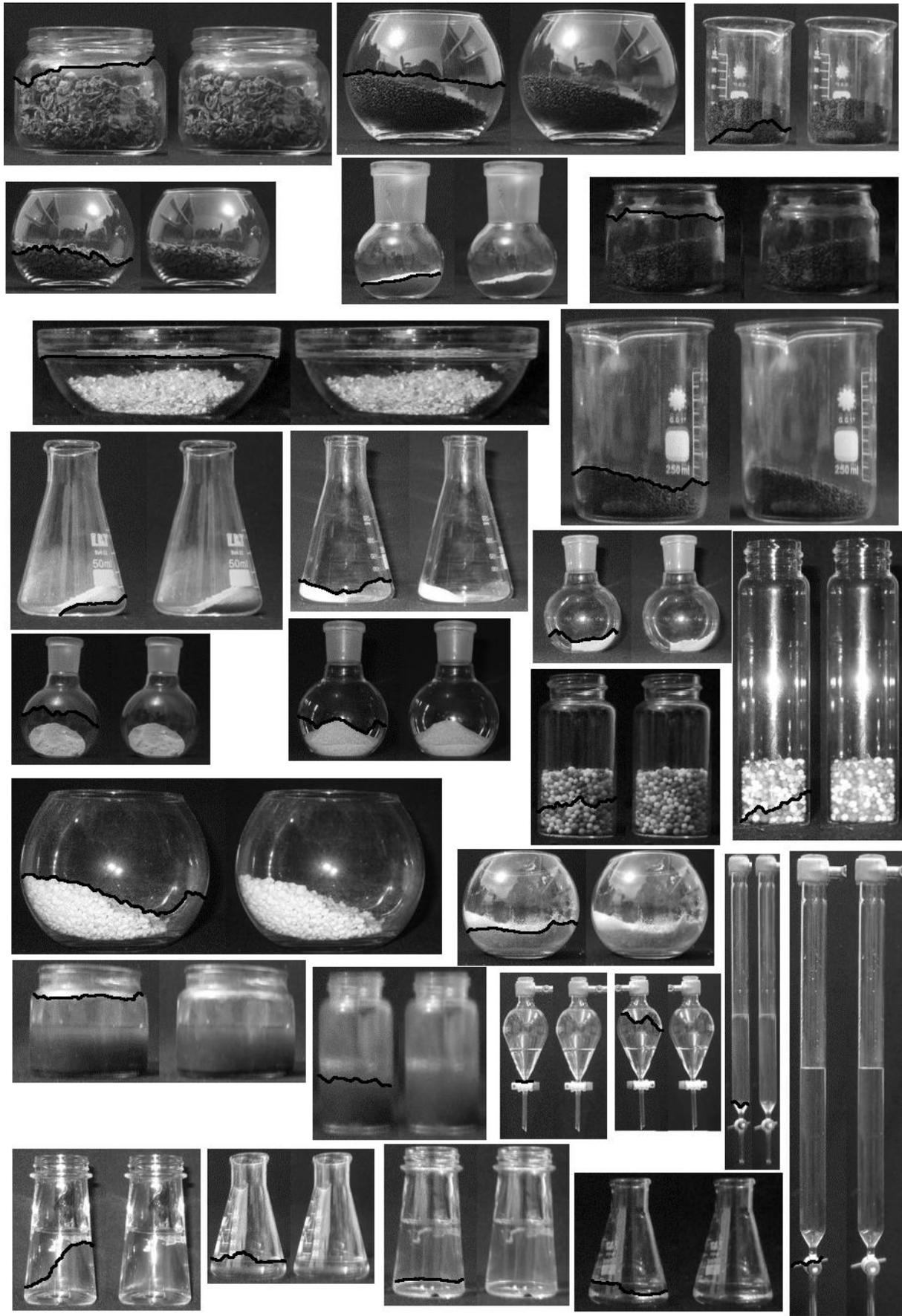

Figure 14. Examples of missed or partial matches between the path found and the material boundary in the image. Based on method 3, Tables 1-2

# 7. Results and discussion

The test results clearly demonstrate the ability of the method to accurately trace the boundaries of the materials in images for various materials and containing vessels. These results can be observed from the high rates of true matches between the traced curve and material boundary in the images, as given in Tables 1-2. Examples for true matches shown in Figure 13. Boundaries of solid materials (Table 1) were traced with higher accuracy compared with those of liquid materials (Table 2), which can be explained by the fact that most of the liquids examined in this work were transparent and therefore have weak edges that were more difficult to trace (Figures 13-14). In comparison, most of the solids used were opaque with strong edges, and thus, their boundaries were easier to find in the image (Figure 13). Use of edges as a material boundary indicator proved superior to use of intensity change across the curve in terms of accuracy, as evidenced by comparing lines 1-2 with lines 3-4 in Tables 1-2. However, all indicators for the phase boundary gave good results, even for materials with blurry boundaries or complex textures[59] (Tables 1-2).

## 7.1. Comparison of edge based indicators for material boundary recognition

Cost functions based on Canny edge detector gave the best result for recognition of materials interface (Tables 1-2). Edge-based boundary indicators that used the difference between edge density on the path and its surroundings gave results superior to methods that used cost based on the path overlap with edges (Section 5.4.1); these results can be observed by comparing lines 3 and 4 in Tables 1-2. The explanation for these results is that indicators based on path overlap with edges are more sensitive to noisy image regions with high edge density (Section 5.4.1).

## 7.2. Comparison of intensity based indicators for material boundary recognition

Cost functions based on intensity change perpendicular to the boundary curve gave good results for liquid and solid material boundaries, although inferior in comparison to edge based indicators (Tables 1-2). Phase boundary indicators based on relative intensity change across the curve (Section 5.3) gave superior results for tracing the boundaries of liquids compared with indicators based on absolute intensity change, although indicators based on absolute intensity change across the path (Section 5.3) produced superior results for solids. These results can be observed by comparing lines 1 and 2 in Tables 1-2. The explanation for this difference is that the liquids in the images examined in this work are transparent and give weaker intensity difference along their boundaries, whereas the solids examined are opaque and give stronger intensity difference (Figure 13).

### 7.3. Main reasons for false recognitions

Various examples of false recognition of material boundaries in the image are shown in Figure 14. The four main causes of false recognitions are:

1) **Weak and blurry material boundaries**: Material interfaces with weak and blurry edges occur mostly in transparent liquids for which the absorption reflection from the liquid is closer to that of the air. Such boundaries show weak intensity change along their border, which make them more difficult to recognize. Blurry boundaries occur mostly for images of liquids with emulsive surfaces in which the transition between phases occurs over a wide region (emulsion layer). As a result, the material image shows a weak intensity gradient over the boundary. Given the importance of sharp changes in intensity for boundary recognitions, this situation makes such surfaces more difficult to trace (Figure 13). Vessels with opaque, dirty, or rigid surfaces are another cause of blurry material boundaries in the images. Previous works[5] showed that use of lower resolution in tracing boundaries of material with blurry/emulsive interfaces could increase their recognition accuracy.

2) **Functional parts of the vessel:** Vessel components such as corks, funnels and valves are a major cause for false recognition (Figure 14). The boundaries of such features often have strong edges that cause them to be identified as the material surface. As explained in Section 4.2.2, accurate definition of the regions of these features in the image and the ability to ignore them in the scan could likely solve this problem.

3) **Patterns on the vessel surface:** Reflections, marks, and other features on the vessel surface are another main cause of false recognitions. The vessel surface can become stained from fragments of the material itself (drops or dust) or from inherited marks on the vessel surface (Figure 14). The false recognition that results from patterns on the vessel surface could be accounted for by comparing or subtracting the image of the filled vessel from an image of the empty vessel or using multiple images of the vessel from different directions.[60,61] Another problem is sharp turns in the vessel surface that cause changes in the reflectance of the vessel surface and induce strong edges. Such regions can be recognized by the corresponding sharp turn on the vessel contour in the image.

4) **Textures and shadow in the material:** Materials with strong textures, i.e., granular materials with particles that have none-uniform colors, are another source of edges that can induce false recognition (Figure 14). In addition, different illumination conditions and shading of different regions of the materials are often caused by turns of the vessel surface and can be another source of patterns that can induce false recognition (Figure 14). Adding texture analysis[43,59] can likely solve this problem.

### 7.4. Percentiles of the cost function as better indicators for boundaries

The overlap of a given path with the material boundary in the image was evaluated based on the path cost, which was calculated as the sum of the costs of all points along this path divided by the path horizontal length (Section 5.2). The path cost is closely related to the average cost of all points on the path. However, previous works[5] have shown that the median/percentile of the cost along a path is a better indicator than the average for evaluating the path correspondences with material boundaries. Unfortunately, Dijkstra's algorithm, which is used for finding the optimal path in this work, can only be used for cases in which the best path is calculated based on the sum of the costs of all path points. Methods for finding the best path based on the percentile/median of local costs offer considerable potential to improve the recognition of phase boundaries.

## 8. Further application of the method

The method described in this work finds the boundary of the material in a transparent vessel using the assumption that this boundary appears in the image as a curve that starts and ends on the contour of the vessel (in addition to certain constraints on the curve slope and path). All of these assumptions are valid not only for the materials inside transparent vessels, but also for the boundary material carried on top of carrier vessels, i.e., plates, spatulas, spoons, bowls and trowels. It is clear that the boundary curve of such materials (viewed from the side) must start and end on the carrier vessel contour line. All previous assumptions with respect to the slope boundary curve slope (Sections 3-4) are valid for this case as well. The only exception is the assumption that the curve must pass entirely inside the vessel boundary (Section 4.2), which is clearly invalid for such cases.

## 9. Summary

This work introduced a general computer vision method for tracing the boundaries of solid and liquid materials in transparent containers. The method is based on the assumption that the boundary of the material inside the vessel must take the shape of a curve that starts and ends on the containing vessel contour. The fact that boundaries and surfaces of real materials have various physical limitations (i.e., limited slope steepness) enabled applications of multiple constraints on the shape and location of the boundary curve in the image. The correspondence of a given curve on the image to a material boundary in this image was evaluated using a cost function, which was calculated based on selected image properties along the curve. The lower the cost of the curve, the higher its probability of matching the real material boundary in the image. A few image properties, i.e., edges and intensity gradients, were examined as indicators for phase boundaries. Based on the above assumption, it was possible to transform the problem of finding the material boundary in the image into a problem of finding the optimal path with minimal cost on a grid. This problem was solved using Dijkstra's algorithm. The method produced good results for tracing the boundaries of solid and liquid materials in a variety of cases, which include different types of vessels and materials. The major sources of errors are blurry boundaries and vessel parts with strong edges, such as corks and funnels. The method can be easily expanded into multiphase systems and identification of material located on top of carrier vessels, i.e., plates, spatulas, bowls and spoons. Further expansion of the method could include use of the image properties of the bulk material (instead of only the image properties along the interface). Development of methods for distinguishing between image patterns resulting from the material boundary and patterns resulting from other features on the vessel surface could also increase accuracy. The method could be applied to automation and validation of various processes in the chemistry lab and in real-world material-handling processes. In addition, the recognition of the material boundaries and region in the image can be used to enable higher level image analysis such as recognition of the material type and properties.[62,63,50,64]

# 10. Appendix

## 10.1. Simplified algorithm

The simplified basic algorithm described in the text (Section 2, Figure 3) is given below in Matlab notation. The remarks are shown in *italic green* fonts. The full code for the method is given as supporting material.

**Input1: Image** *%Image of the vessel containing the material*
**Input2: VesselContourPoints** *% Array containing pixels corresponding to vessel boundary in Image*
**Input3: NumVesselContourPoints** *% Number of points in VesselContourPoints array*
-----------------------------------------------------------------------------------------------------------------------------------
**For f=1:NumVesselContourPoints** *%Scan all points on the vessel contour*
   **Endpoint1=VesselContourPoints(f);** *% Define first endpoint of the path*
   **For f2=f1+1:NumVesselContourPoints** *%Scan all points on the vessel contour*
     **Endpoint2=VesselContourPoints(f2);** *%Define second endpoint of the path*
     **LegalPathRegion=FindLegalPropogationRegion(EndPoint1,EndPoint2);**
*% Find the legal region in the image in which the path between the two endpoints can pass*
     **(Path,PathCost)= FindBestPath(EndPoint1,EndPoint2, LegalPathRegion,Image);**
*%Find the best (Cheapest) path on Image between the two endpoints that pass only through the legal region*
       **If (PathCost<MinimalPathCost)**
         **MinimalPathCost=PathCost;**
         **MaterialBoundary= Path;** *%Boundary of the material in the image*
       **end;**
*%If the cost of this path is lower than the cost of any previous path, use this path as the new material boundary*
   **end;** *%End of loop with f1*
**end;** *%End of loop with f2*
-----------------------------------------------------------------------------------------------------------------------------------
**Output: MaterialBoundary** *%Boundary of the material in the image*

## 10.2. Algorithm fast version

The simplistic algorithm, described in appendix 10.1, demands the calculation of the optimal path for each pair of points on the vessel contour separately and is therefore time consuming. A better method is to calculate all the optimal paths leading from a given starting point $E$ in one round. This is indeed possible using Dijkstra's algorithm. This, however, leads to a problem: The legal region in which the path is allowed to pass is defined by both endpoints of the path (Figures 7b,c). Hence, if only the starting point ($S$, Figure 15a) is known, the resulting path might pass in image regions that are illegal for the second endpoint ($E$, Figure 15b). This means that calculating all optimal paths from point $S$ might result in the optimal path from $S$ to $E$ passing illegal image regions with respect to point $E$ (Figure 15). If this is the case, then the optimal path from $S$ to $E$ will have to be recalculated. However, given that the method is only looking for the single lowest cost path in the entire image, this will need to happen only if the path cost is lower than the cost of any previously calculated path, which is a rare event. Hence, this will happen in very few cases and will not significantly slow the program.

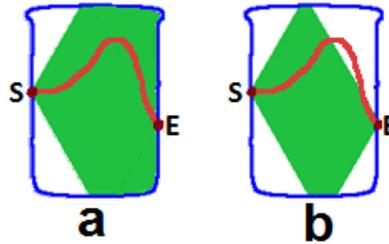

**Figure 15: A path pass in legal regions with respect to endpoint S** *(a)* **might pass in an illegal region with respect to end point E (b). Legal region marked green path marked red.**

A version of the algorithm based on the above concept is given below in Matlab notation. The remarks are shown in *italic green* fonts. The full code for the method is given as supporting material.

```
Input1: Image %Image of the vessel containing the material
Input2: VesselContourPoints % Array containing pixels corresponding to the vessel boundary in Image
Input3: NumVesselContourPoints % Number of points in VesselContourPoints array
-----------------------------------------------------------------------------------------------------------------------------
For f=1:NumVesselContourPoints  %Scan all points on the vessel contour
    Endpoint1=VesselContourPoints(f);  % Define first endpoint of the path
        LegalPathRegion1=FindLegalPropogationRegionPoint1(EndPoint1);
% Find the legal region in the image in which the paths leading from endpoint1 can pass.
        (Path,PathCostMatrix)= FindAllBestPaths(EndPoint1, LegalPathRegion1,Image);
%Find all the best path leading from EndPoint1 that pass only through the legal regions of  EndPoint1.
%Return a matrix (PathCostMatrix) of the cost of the path from EndPoint1 to every point on the image
        For f2=f1+1:NumVesselContourPoints  %Scan all points on the vessel contour
        Endpoint2=VesselContourPoints(f2);  %Define second endpoint of the path
            If (PathCostMatrix(EndPoint2)<MinimalPathCost)
%If the cost of this path from EndPoint1 to EndPoint2 is lower than the cost of any previous path found, then
%consider this path as the new material boundary
                LegalPathRegion12=FindLegalPropogationRegion12(EndPoint1,EndPoint2);
% Find the legal region in the image in which the path between the two endpoints can pass
                If (Path inside LegalPathRegion)
%if the path found pass only inside legal region defined for both endpoints write it as a new best path
                    MinimalPathCost=PathCost;
                    MaterialBoundary=PathMat(EndPoint2); %Boundary of the material in the image
                else
%If the path pass in an illegal region with respect to the endpoint2, recalculate the path between  endpoints
                    (Path,PathCost)= FindBestPath(EndPoint1,EndPoint2, LegalPathRegion,Image);
%Find the best (Cheapest) path on Image between the two endpoints that pass only through the legal region
%for both endpoints
                    If (PathCost<MinimalPathCost)
                        MinimalPathCost=PathCost;
                        MaterialBoundary= Path; %Boundary of the material in the image
                    end;
%If the cost of this path is lower than the cost of any previous path use this path as the new material boundary
            end;
        end; %End of loop with f1
 end; %End of loop with f2
-----------------------------------------------------------------------------------------------------------------------------
Output: MaterialBoundary %Boundary of the material in the image
```

## 11. Supporting materials.
Code and documentation for the method described is freely available at:

http://www.mathworks.com/matlabcentral/fileexchange/49076-find-the-boundaries-of-materials-in-transparent-vessels-using-computer-vision

Code and documentation for method for finding the boundary of the vessel containing the material in the image (necessary as input to the method) is freely available at:

http://www.mathworks.com/matlabcentral/fileexchange/46887-find-boundary-of-symmetric-object-in-image

And

http://www.mathworks.com/matlabcentral/fileexchange/46907-find-object-in-image-using-template--variable-image-to-template-size-ratio-